\documentclass[conference]{IEEEtran}

\usepackage[T1]{fontenc}
\usepackage{graphicx}
\usepackage{booktabs}
\usepackage{amsmath,amssymb}
\usepackage{microtype}
\usepackage{url}
\usepackage[hidelinks]{hyperref}
\usepackage{flushend}
\usepackage{placeins}
\usepackage{tikz}
\usetikzlibrary{positioning,arrows.meta}

\hypersetup{
    pdftitle={Diagnosing Zero-Shot Forgetting in CLIP Fine-Tuning: Attention Drift Characterizes Adaptation, Representation Drift Predicts Transfer Degradation},
    pdfauthor={Ruize Xia},
    pdfkeywords={CLIP, attention, fine-tuning, LoRA, zero-shot transfer, vision transformers}
}

\title{Diagnosing Zero-Shot Forgetting in CLIP Fine-Tuning:\\
Attention Drift Characterizes Adaptation, Representation Drift Predicts Transfer Degradation}

\author{\IEEEauthorblockN{Ruize Xia}
\IEEEauthorblockA{\textit{High School Student}\\
Nanjing Foreign Language School\\
Nanjing, China\\
xiaruize0911@gmail.com}}

\begin{document}
\maketitle

\begin{abstract}
Fine-tuning CLIP on a downstream task typically reduces its zero-shot accuracy on other tasks, and quantifying this loss ordinarily requires a labeled transfer benchmark. We study whether the loss can be anticipated from unlabeled target images that are already available during fine-tuning. Four signals are compared: the entropy of CLS token attention, linear centered kernel alignment (CKA) with the pretrained model, embedding drift, and relative weight change. ViT-B/32 is adapted by full fine-tuning (Full FT) and by LoRA on EuroSAT and Oxford-IIIT Pets over four learning rates and three seeds, with an additional ViT-B/16 grid; DTD and GTSRB are held out as confirmation targets, and forgetting is measured as zero-shot retention on CIFAR-100, Food-101, and ImageNet-R. Within a fixed configuration, the change in attention entropy is strongly associated with forgetting (mean $|\rho|=0.860$). Its sign, however, depends on the dataset, the adaptation method, and the backbone, and the pooled correlation falls to 0.361. A block decomposition shows that this sign is determined almost entirely by blocks 7 to 12. CKA ranks runs by retention at $\rho=0.960$, and the ranking is already established after the first of six epochs ($\rho=0.959$). On the held-out targets, CKA reaches $\rho=0.95$, and the block account correctly predicts the sign of the attention change for both datasets. These results support a simple decision rule: CKA is suited to comparisons across configurations, whereas attention entropy is informative only within a single configuration. The evidence is limited to one model family and short training schedules.
\end{abstract}


\section{Introduction}
CLIP learns visual representations that transfer to many recognition tasks without further training. Fine-tuning on a specific task usually improves accuracy on that task at the cost of this broader capability. The cost is normally measured only after training, on a separate labeled benchmark, which may be expensive to construct or unavailable in specialized domains.

This paper examines whether the loss of zero-shot capability can be detected during fine-tuning from internal model changes, measured only on unlabeled target images that are already at hand. Our initial hypothesis was that CLS token attention becomes more concentrated under aggressive fine-tuning and that this concentration tracks forgetting. We evaluate the hypothesis against alternatives based on representations and weights and find that it holds only in part. The work is an empirical study of diagnostic signals rather than a proposal for a new fine-tuning method. Its contributions are as follows:
\begin{enumerate}
    \item We show that attention entropy characterizes the magnitude of adaptation rather than the resulting transfer loss. It is reliable within one configuration but changes sign across configurations, and we trace the sign to the late blocks (Sec.~\ref{sec:why-sign}).
    \item We show that CKA with the pretrained model is the most accurate ranking signal across configurations and retains a consistent direction in every group. Its predictive value after one epoch is explained by the early stabilization of the ordering of runs (Sec.~\ref{sec:why-early}), and both findings are confirmed on two held-out target datasets and two additional transfer benchmarks (Sec.~\ref{sec:heldout}).
    \item We confirm that, at comparable target accuracy, LoRA retains substantially more zero-shot accuracy than Full FT, and that weight interpolation moves models along the resulting accuracy and retention frontier.
    \item We derive a practical decision procedure (Fig.~\ref{fig:pipeline}) that specifies which signal to monitor, when to measure it, and how to act on it.
\end{enumerate}

\section{Related Work}
\textbf{Fine-tuning CLIP.}
CLIP \cite{radford2021learning} learns image representations from natural language supervision. Kumar et al.\ show that conventional fine-tuning can distort pretrained features and degrade performance out of distribution \cite{kumar2022finetuning}. Proposed remedies include weight interpolation between the zero-shot and fine-tuned models \cite{wortsman2022robust}, task arithmetic \cite{ilharco2023editing}, fine-tuning with the pretraining objective \cite{goyal2023finetune}, LoRA \cite{hu2022lora}, and residual feature adapters \cite{gao2024clipadapter}. These methods are evaluated by final accuracy on held-out benchmarks such as ImageNet-R \cite{hendrycks2021many}. We instead study signals that can be measured during training without labels.

\textbf{Attention and representation similarity.}
Vision transformers compute attention over image patches \cite{dosovitskiy2021image}. Because attention weights are not causal explanations \cite{jain2019attention,wiegreffe2019attention}, we treat them only as structural measurements. Zhai et al.\ associate the collapse of attention entropy with unstable transformer training \cite{zhai2023stabilizing}; we ask whether the same quantity tracks forgetting during adaptation. Since concentrated attention can also be benign in trained models \cite{xiao2024efficient,darcet2024vision}, all attention statistics are expressed relative to the pretrained model. CKA measures the similarity of network representations \cite{kornblith2019similarity}, and the projection norm relates update magnitude to error out of distribution \cite{yu2022predicting}.

\section{Method}
\subsection{Diagnostic Signals}
All signals are computed on a fixed \emph{probe set} of a few hundred unlabeled target images and are expressed relative to the pretrained model. Table~\ref{tab:signals} summarizes them.

For block $\ell$ and head $h$, let $a_{\ell h}$ denote the attention of the CLS token over the $N_p$ patch tokens, renormalized after removing the CLS token's attention to itself ($N_p=49$ for ViT-B/32 and 196 for ViT-B/16). Its entropy, in nats, is
\begin{equation}
H(a_{\ell h})=-\sum_{j=1}^{N_p}a_{\ell hj}\log a_{\ell hj},
\end{equation}
averaged over images, heads, and the 12 blocks. For any quantity $M$, the relative change from the pretrained model is
\begin{equation}
\Delta M = 100\,\frac{M_{\mathrm{adapted}}-M_{\mathrm{pretrained}}}{M_{\mathrm{pretrained}}}.
\end{equation}
A positive $\Delta H$ indicates that attention is distributed over more patches, and a negative value indicates concentration. We refer to $\Delta H$ as the \emph{signed} change and to $|\Delta H|$ as its magnitude. Because an average over blocks can conceal changes of opposite sign, we also report the last block, $\Delta H_{12}$.

\begin{table}[!t]
\caption{Signals compared. All are measured relative to pretrained CLIP. None requires labels, and weight change also requires no images.}
\label{tab:signals}
\centering
\scriptsize
\setlength{\tabcolsep}{3pt}
\begin{tabular}{lll}
\toprule
Signal & Definition & Input \\
\midrule
$\Delta H$ & change in CLS token attention entropy, Eqs.~(1) and (2) & attention \\
CKA & linear CKA of CLS token features, mean over 12 blocks & features \\
$D_z$ & $1-\cos(z_0,z)$ between output embeddings & features \\
$D_W$ & $\lVert\Delta W\rVert_2/\lVert W_0\rVert_2$ over encoder weights & weights \\
\midrule
Gini, ERF & attention concentration (auxiliary) & attention \\
\bottomrule
\end{tabular}
\end{table}

CKA is computed between the CLS token features of the pretrained and adapted models: the features of each block are centered over the probe images, and the normalized Hilbert\textendash Schmidt inner product of the resulting Gram matrices is averaged over blocks \cite{kornblith2019similarity}. A value of 1 indicates no change. We also report the mean over blocks 1 to 4 (early CKA) and block 12 alone. For LoRA, $\Delta W=\frac{\alpha}{r}BA$, which places both methods on a common scale. Training loss and target accuracy serve as baselines; only the latter requires labels. ERF@0.95 is the fraction of patches required to accumulate 95\% of the attention mass.

\subsection{Experimental Protocol}
We fine-tune CLIP ViT-B/32 on EuroSAT \cite{helber2019eurosat} and Oxford-IIIT Pets \cite{parkhi2012cats} using training subsets, balanced across classes, of 1{,}600 and 1{,}480 images. Validation sets (300 and 370 images) and probe sets (200 and 222 images) are drawn from the training split and are disjoint from the training subset. Checkpoints are selected by validation accuracy, and the test split is evaluated once per run, so no reported accuracy is used for model selection. Training uses batch size 32, six epochs, AdamW with weight decay 0.01, a 5\% linear warmup followed by cosine decay, and gradient clipping at 1.0. The learning rates are $3\cdot 10^{-6}$, $10^{-5}$, $3\cdot 10^{-5}$ and $10^{-4}$, and the seeds are 7, 11 and 19. The text encoder is frozen throughout. Full FT updates the image encoder, its projection, and a linear classifier. LoRA inserts rank 8 updates ($\alpha=16$, dropout 0.05) into every query and value projection and also trains the projection and classifier. Reference configurations comprise a linear probe, tuning of the last block only, and LoRA with a frozen projection. Repeating the grid at three learning rates on ViT-B/16 adds 36 runs. After the analysis had been fixed on EuroSAT and Pets, we added DTD \cite{cimpoi2014describing} (47 texture classes) and GTSRB \cite{stallkamp2012man} (43 traffic sign classes) as held-out targets, with the same grid and 40 training images per class (48 runs). They are used only to test the conclusions. Each run records all signals at every epoch, and a single script regenerates all tables and figures.\footnote{\href{https://github.com/xiaruize0911/Attention_Collapse_in_CLIP_Fine-tuning_repo}{github.com/xiaruize0911/\allowbreak Attention\_Collapse\_in\_CLIP\_Fine-tuning\_repo}.}

\subsection{Measuring Forgetting}
Zero-shot accuracy is computed with the adapted image encoder, including any LoRA weights, against text embeddings of ``a photo of a \{class\}.'' This detail matters: evaluating a LoRA checkpoint through an unmodified CLIP model would measure the pretrained model instead. \emph{Retention} is the percentage of pretrained zero-shot accuracy that is preserved. The primary outcome is CIFAR-100 \cite{krizhevsky2009learning} (all 10{,}000 test images; pretrained accuracy 61.33\%). CIFAR-10 produces an almost identical ranking of runs ($\rho=0.971$) and is therefore not treated as an independent outcome. As secondary outcomes, class balanced subsets of Food-101 \cite{bossard2014food} (1{,}919 images) and ImageNet-R \cite{hendrycks2021many} (2{,}000 images) are scored for the held-out runs and for a GPU rerun of the 48 core runs, whose CIFAR-100 retention agrees with the original at $\rho=0.998$.

Each signal is evaluated by (i) its Spearman correlation with final retention over the 48 core runs (Full FT and LoRA on ViT-B/32), (ii) the same correlation computed from its value after the first epoch, and (iii) its selection utility, defined as the retention of the run it selects among runs within two points of the best validation accuracy. Correlations are computed over runs; bootstrap intervals resample the 16 configuration means, since seeds of the same configuration are not independent.

\begin{figure}[!t]
    \centering
    \includegraphics[width=\columnwidth]{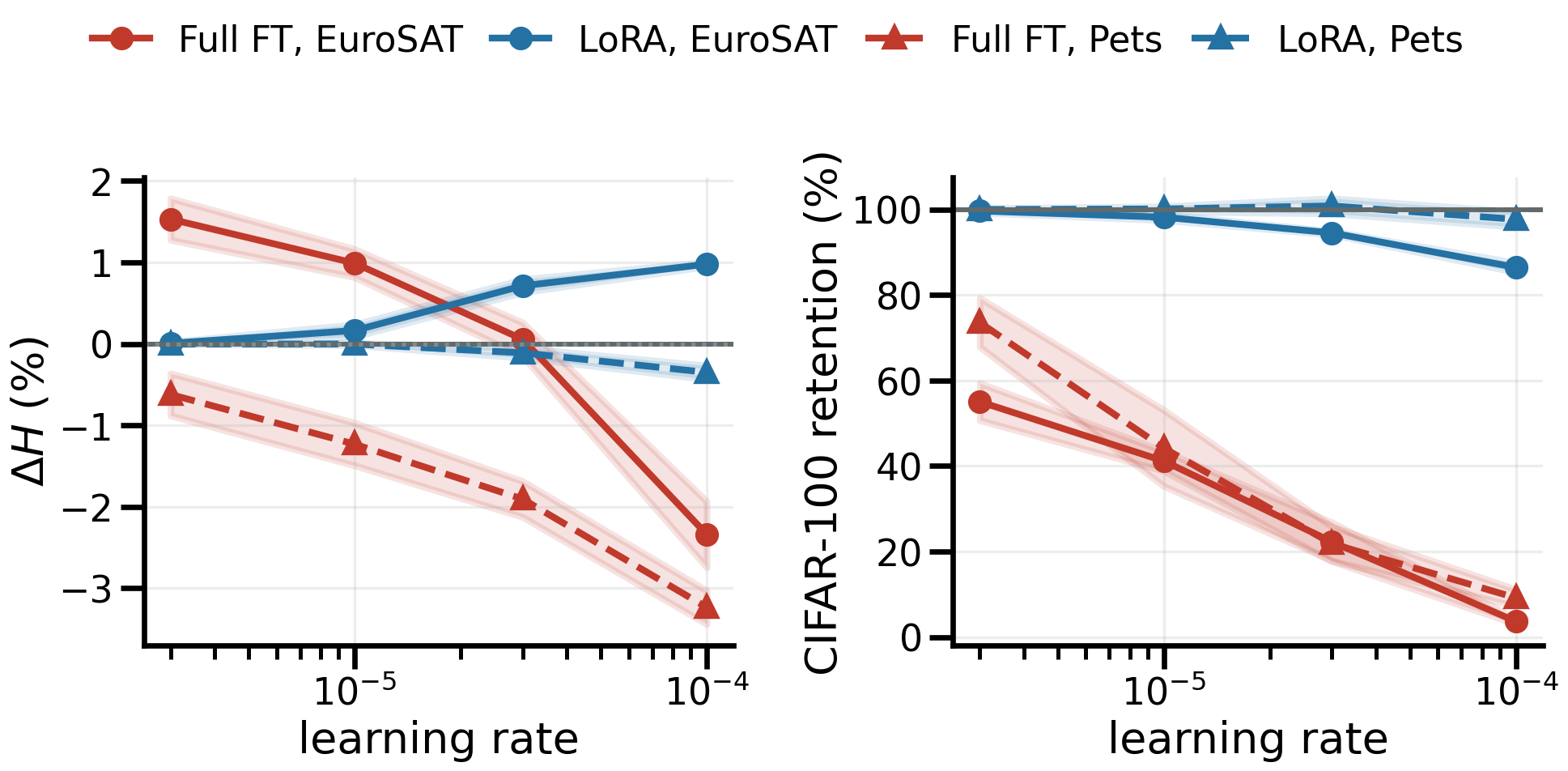}
    \caption{Effect of learning rate on the primary targets (mean and standard deviation over three seeds; grey: linear probe). Full FT and LoRA change attention in opposite directions on EuroSAT, while retention declines for both except LoRA on Pets.}
    \label{fig:dose}
\end{figure}

\begin{table*}[!t]
\caption{Configurations selected on validation accuracy (mean $\pm$ s.d.\ over three seeds). Retention is the percentage of pretrained zero-shot accuracy preserved; for EuroSAT and Pets, Food-101 and ImageNet-R come from the GPU rerun. $^{\ddagger}$Held-out targets. The full grid is released with the code.}
\label{tab:grid}
\centering
\scriptsize
\setlength{\tabcolsep}{4pt}
\renewcommand{\arraystretch}{0.9}
\begin{tabular}{llrrrrrrr}
\toprule
 & & & & & & \multicolumn{3}{c}{Retention (\%)} \\
\cmidrule(lr){7-9}
Dataset & Method & Learning rate & Test acc. (\%) & $\Delta H$ (\%) & CKA & CIFAR-100 & Food-101 & ImageNet-R \\
\midrule
EuroSAT & Full FT & $10^{-5}$ & 97.2\,$\pm$\,0.3 & 0.99\,$\pm$\,0.17 & 0.802\,$\pm$\,0.015 & 41.2\,$\pm$\,2.3 & 83.6 & 73.7 \\
EuroSAT & LoRA r=8 & $10^{-4}$ & 95.2\,$\pm$\,0.4 & 0.98\,$\pm$\,0.03 & 0.900\,$\pm$\,0.004 & 86.4\,$\pm$\,1.1 & 99.0 & 97.2 \\
Pets & Full FT & $3\cdot 10^{-5}$ & 88.3\,$\pm$\,0.3 & -1.90\,$\pm$\,0.22 & 0.676\,$\pm$\,0.024 & 21.8\,$\pm$\,4.0 & 39.7 & 29.4 \\
Pets & LoRA r=8 & $10^{-4}$ & 89.9\,$\pm$\,0.5 & -0.35\,$\pm$\,0.08 & 0.935\,$\pm$\,0.004 & 97.7\,$\pm$\,1.8 & 97.8 & 96.6 \\
\midrule
DTD$^{\ddagger}$ & Full FT & $10^{-5}$ & 71.8\,$\pm$\,1.2 & 1.87\,$\pm$\,0.02 & 0.743\,$\pm$\,0.005 & 21.4\,$\pm$\,5.2 & 33.8 & 38.4 \\
DTD$^{\ddagger}$ & LoRA r=8 & $10^{-4}$ & 71.2\,$\pm$\,0.8 & 0.86\,$\pm$\,0.05 & 0.945\,$\pm$\,0.004 & 85.0\,$\pm$\,1.0 & 87.4 & 85.5 \\
GTSRB$^{\ddagger}$ & Full FT & $3\cdot 10^{-5}$ & 94.6\,$\pm$\,0.8 & -5.96\,$\pm$\,0.37 & 0.711\,$\pm$\,0.007 & 7.0\,$\pm$\,1.4 & 8.8 & 19.7 \\
GTSRB$^{\ddagger}$ & LoRA r=8 & $10^{-4}$ & 80.8\,$\pm$\,2.8 & -2.52\,$\pm$\,0.10 & 0.876\,$\pm$\,0.002 & 67.2\,$\pm$\,1.8 & 97.6 & 95.7 \\
\bottomrule
\end{tabular}
\end{table*}

\section{Results}
\subsection{Learning Rate Governs Drift and Forgetting}
Fig.~\ref{fig:dose} shows that larger learning rates produce larger internal changes and more forgetting. Retention decreases monotonically with learning rate in three of the four primary groups, for example from 55.0\% to 3.7\% for EuroSAT Full FT. Pets LoRA retains between 97.7\% and 100.8\% throughout, which leaves too little variation for rank correlations to be interpreted. With four learning rates per group, these trends are descriptive and are not tested for significance.

The direction of the attention change is less regular. On EuroSAT, Full FT broadens attention at the smallest rate (1.53\,$\pm$\,0.25\%) and concentrates it at the largest (-2.34\%), whereas LoRA broadens it increasingly; on Pets, both methods concentrate it. At the configurations selected on EuroSAT validation, Full FT and LoRA exhibit nearly identical $\Delta H$ (+0.99\% and +0.98\%) but retain 41.2\% and 86.4\% of CIFAR-100 accuracy.

\textbf{LoRA forgets less at comparable accuracy.} Table~\ref{tab:grid} compares the configurations selected on validation. On EuroSAT, Full FT reaches 97.2\% target accuracy with 41.2\% retention, and LoRA reaches 95.2\% with 86.4\%. On Pets and on the held-out DTD, the two methods are within two points in accuracy, and LoRA retains far more. On GTSRB, LoRA is less accurate (80.8\% vs.\ 94.6\%) but retains 67.2\% rather than 7.0\%. The ordering is the same on Food-101 and ImageNet-R and is insensitive to an ensemble of 7 prompts and to LoRA ranks of 4 and 16.

\subsection{Predictive Validity of the Signals}
Across the 48 core runs, CKA ranks runs by retention at $\rho=0.960$ and embedding drift at $\rho=-0.919$, whereas signed $\Delta H$, which motivated this study, reaches only 0.361; Gini and ERF reach $|\rho|=0.844$ and $0.796$. The CKA result survives three checks: excluding configurations with mean target accuracy below 50\%, in which the encoder barely changes ($\rho=0.962$); restricting CKA to blocks 1 to 4 (0.921), so the association is not confined to the output layer; and comparison with a labeled subset of 1{,}000 CIFAR-100 images, which reaches 0.967 after one epoch but requires labels from the benchmark being predicted and is therefore an upper bound. With 16 configurations, the bootstrap separates CKA from signed $\Delta H$ (difference +0.601, 95\% CI [0.149, 0.948]) but not from embedding drift or weight change.

\begin{figure}[!t]
    \centering
    \includegraphics[width=0.9\columnwidth]{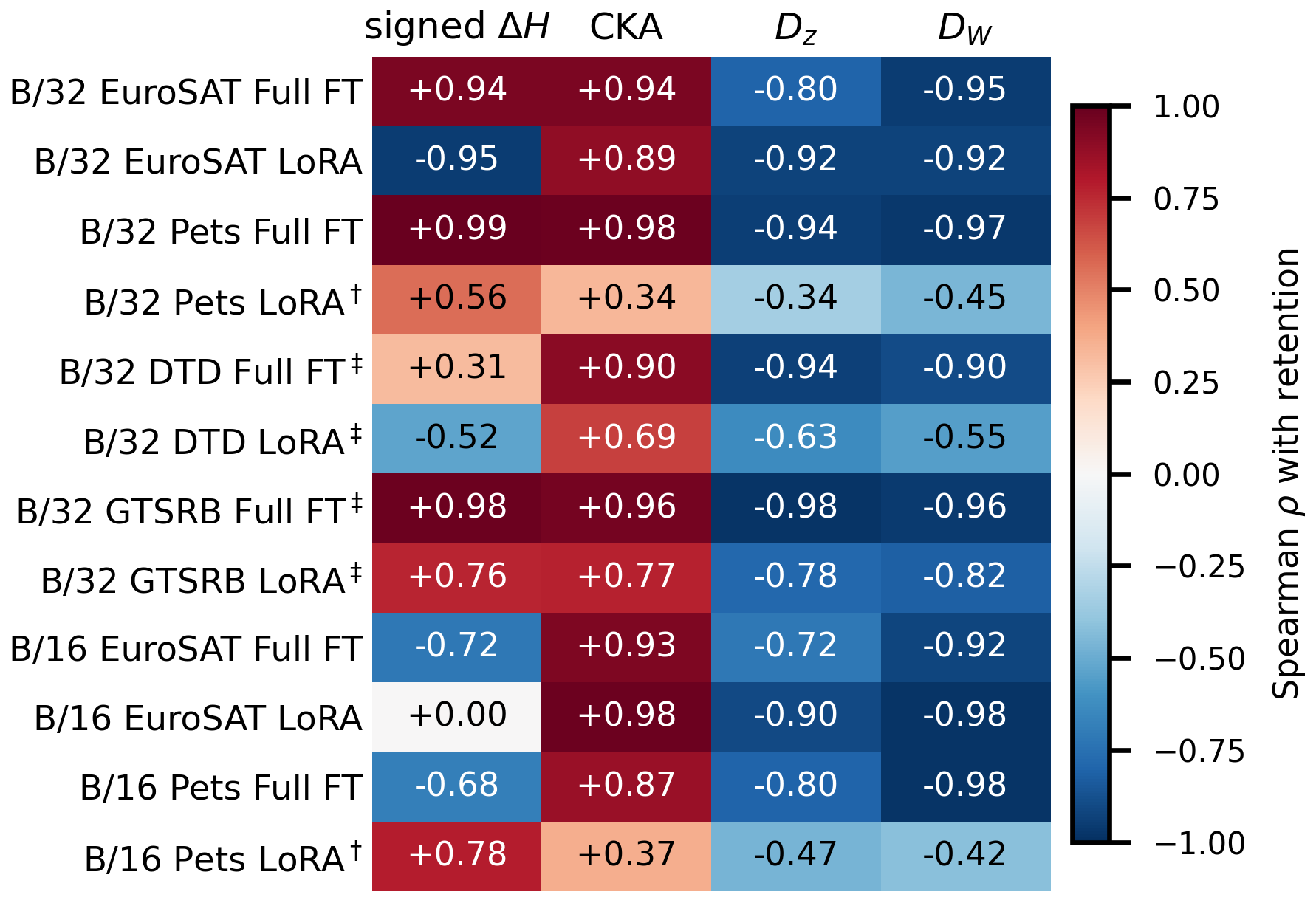}
    \caption{Spearman correlation of each signal with CIFAR-100 retention within each backbone, dataset, and method group. Signed $\Delta H$ changes sign between groups; CKA, $D_z$, and $D_W$ do not. $^{\ddagger}$Held-out target. $^{\dagger}$Retention varies by less than 6 points, so $\rho$ is not interpreted.}
    \label{fig:synthesis}
\end{figure}

Within a single dataset and method, the ordering differs. Signed $\Delta H$ attains the highest mean $|\rho|$ (0.860), but only because the absolute value is taken after the sign is observed. When the sign must be fixed in advance, its mean correlation falls to +0.385, whereas CKA retains +0.787. Fig.~\ref{fig:synthesis} shows the same pattern across backbones: $\Delta H$ changes sign between groups, while CKA, embedding drift, and weight change maintain a consistent direction. This instability affects selection. Among EuroSAT runs within two points of the best validation accuracy, $|\Delta H|$ selects a run with 24.0\% retention, below the 41.2\% expected under random selection, whereas embedding drift and CKA select runs with 56.9\% and 50.1\%.

\begin{table}[!t]
\caption{Pooled Spearman correlation of each signal with retention. Primary: 48 EuroSAT and Pets runs; held out: 48 DTD and GTSRB runs; E1: signal after the first epoch.}
\label{tab:heldout}
\centering
\scriptsize
\setlength{\tabcolsep}{2.6pt}
\begin{tabular}{lrrrrrrr}
\toprule
 & \multicolumn{3}{c}{Primary targets} & \multicolumn{4}{c}{Held-out targets} \\
\cmidrule(lr){2-4}\cmidrule(lr){5-8}
Signal & C100 & Food & IN-R & C100 & Food & IN-R & C100 (E1) \\
\midrule
Signed $\Delta H$ & +0.36 & +0.47 & +0.44 & +0.31 & +0.19 & +0.11 & +0.10 \\
$|\Delta H|$ & -0.84 & -0.75 & -0.82 & -0.73 & -0.69 & -0.64 & -0.94 \\
CKA & +0.96 & +0.91 & +0.94 & +0.95 & +0.93 & +0.89 & +0.94 \\
Early CKA & +0.92 & +0.92 & +0.94 & +0.91 & +0.91 & +0.91 & -- \\
$D_z$ & -0.92 & -0.89 & -0.91 & -0.94 & -0.92 & -0.88 & -0.95 \\
$D_W$ & -0.84 & -0.82 & -0.85 & -0.88 & -0.86 & -0.83 & -0.94 \\
\bottomrule
\end{tabular}
\end{table}

\begin{figure*}[!t]
    \centering
    \includegraphics[width=\textwidth]{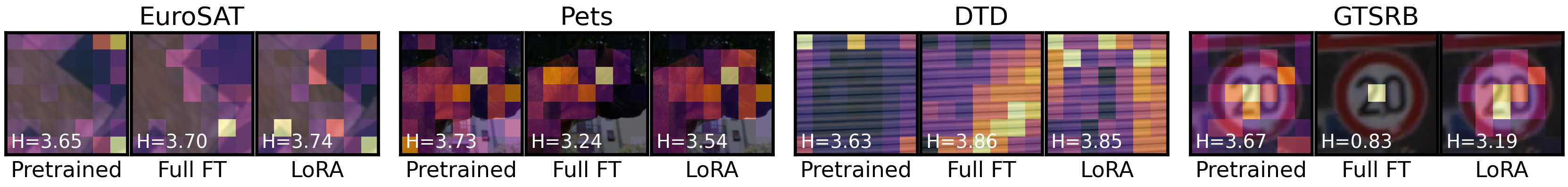}
    \caption{Last block CLS attention on one probe image per dataset for pretrained CLIP, Full FT, and LoRA at $10^{-4}$ (seed 7), with entropy $H$ in nats. Attention broadens on the DTD texture and concentrates on the Pets and GTSRB objects, less so under LoRA. Single images are illustrative only.}
    \label{fig:maps}
\end{figure*}

\subsection{Confirmation on Held-out Targets and Outcomes}
\label{sec:heldout}
Table~\ref{tab:heldout} repeats the evaluation on the held-out targets and the secondary outcomes, and the ordering of signals is unchanged. On DTD and GTSRB, CKA ranks runs by CIFAR-100 retention at $\rho=0.952$ and embedding drift at $-0.945$, whereas signed $\Delta H$ reaches only 0.311. CKA measured after the first epoch still reaches 0.938. Within groups, the correlation of CKA with retention is positive in all 4 held-out groups, whereas that of signed $\Delta H$ is positive in 3 of 4 (Fig.~\ref{fig:synthesis}). At the configurations selected on DTD, Full FT again has the larger entropy change (+1.87\% vs.\ +0.86\%) and far lower retention. One result does not replicate: on DTD, $|\Delta H|$ selects the same run as CKA, so its selection failure on EuroSAT is not general.

\begin{figure}[!t]
    \centering
    \includegraphics[width=\columnwidth]{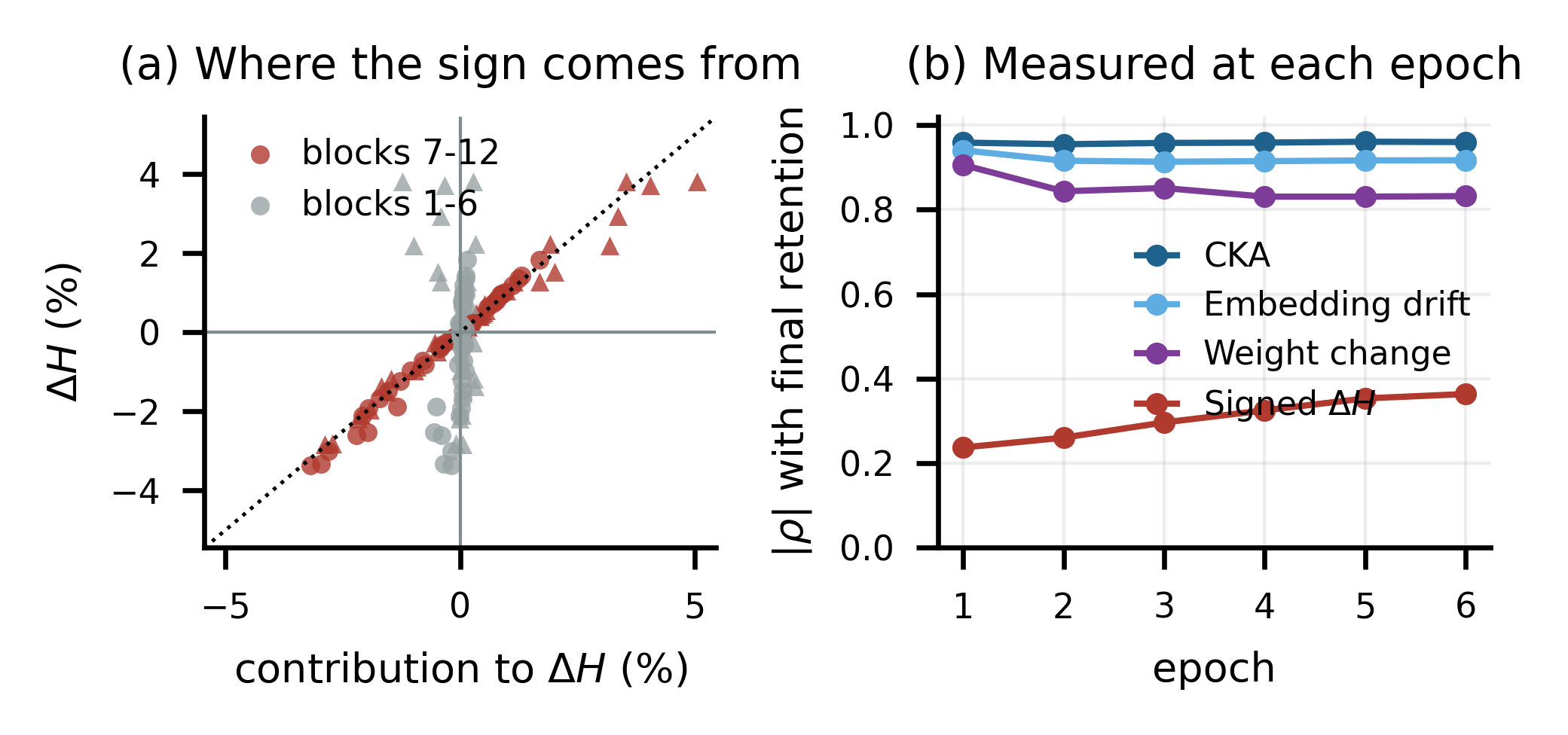}
    \caption{(a) Signed $\Delta H$ of each run against the contribution of blocks 7 to 12 (red) and 1 to 6 (grey); circles: ViT-B/32, triangles: ViT-B/16. The late blocks lie on the diagonal and determine the sign. (b) $|\rho|$ between each signal after epoch $e$ and final CIFAR-100 retention (48 runs).}
    \label{fig:mechanism}
\end{figure}

\subsection{Origin of the Sign of Attention Drift}
\label{sec:why-sign}
Averaged over blocks, $\Delta H$ is an exact sum of the changes in individual blocks, so it can be decomposed into contributions from blocks 1 to 6 and from blocks 7 to 12. The late contribution determines the sign (Fig.~\ref{fig:mechanism}a): its sign agrees with that of $\Delta H$ in all 48 ViT-B/32 runs and all 36 ViT-B/16 runs, and it accounts for a median of 93\% of the total change. The early contribution agrees in only 27 of 48 runs, which is close to chance. Even under Full FT at $10^{-4}$, blocks 1 to 4 remain within 1.8\% of the pretrained values, while block 12 changes by -18.5\% on Pets.

Why do the late blocks broaden attention in some configurations and concentrate it in others? On ViT-B/32, the late contribution is positive in 7 of 8 EuroSAT configurations and in 0 of 8 Pets configurations; on ViT-B/16 the separation is weaker (5 of 6 and 3 of 6). We interpret the direction as reflecting the spatial distribution of task evidence: land use classes are defined by texture across the whole tile, whereas pet breeds are determined by a single object; large updates can override this, as EuroSAT Full FT at $10^{-4}$ shows. This interpretation was stated before the held-out runs and predicts broadening on the texture dataset DTD and concentration on GTSRB, where a single sign determines the class. The late contribution is positive in 7 of 8 DTD configurations and in 0 of 8 GTSRB configurations, and its sign again matches that of $\Delta H$ in all 16 held-out configurations. Fig.~\ref{fig:maps} illustrates both directions; under Full FT, attention on a traffic sign collapses onto one patch. The sign thus reflects the task and the step size rather than the extent of forgetting, which explains why $\Delta H$ tracks the learning rate within a configuration but cannot be compared across configurations.

\subsection{Early Predictive Value of CKA}
\label{sec:why-early}
Fig.~\ref{fig:mechanism}(b) reports the correlation of each signal with final retention at every epoch. CKA measured after the first epoch predicts final retention as well as CKA measured at the end of training ($\rho=0.959$ vs.\ 0.960), because the ordering of runs by CKA changes little after that point (Spearman correlation of 0.998 between epochs 1 and 6). Under a short warmup and cosine decay, the first epoch is trained near the peak learning rate, and the median run completes 40\% of its final weight change within this first sixth of training. Signed $\Delta H$ behaves differently: its correlation increases gradually from 0.24 to 0.36, and 5 runs change sign during training. On the held-out targets, the ordering by CKA is similarly stable (Spearman 0.976 between epochs 1 and 6). This account is empirical and specific to our schedule; with a longer warmup, the first epoch may be less informative.

\begin{figure}[!t]
\centering
\begin{tikzpicture}[
    font=\scriptsize,
    box/.style={draw, rounded corners=1.5pt, align=center, inner sep=2.2pt, minimum height=6mm},
    ask/.style={draw, align=center, inner sep=2.2pt, fill=gray!10},
    go/.style={box, fill=blue!7},
    alt/.style={box, fill=red!7},
    arr/.style={-{Stealth[length=1.6mm]}, semithick}]
\node[box] (probe) at (-29mm, 0) {unlabeled target\\images (probe set)};
\node[box] (train) at (-5mm, 0) {fine-tune\\one epoch};
\node[box] (measure) at (23mm, 0) {measure CKA, $D_z$, $\Delta H$\\against pretrained CLIP};
\node[ask] (q) at (0, -12mm) {comparing different methods,\\datasets, or learning rates?};
\node[go] (cka) at (-23mm, -24mm) {rank candidates by CKA\\or $D_z$ (lower CKA: more loss)};
\node[alt] (dh) at (23mm, -24mm) {one fixed setting: follow\\$\Delta H$ once its sign is known};
\node[box] (act) at (0, -36mm) {if much forgetting is expected: lower the rate,\\switch to LoRA, or interpolate toward $\theta_0$};
\draw[arr] (probe) -- (train);
\draw[arr] (train) -- (measure);
\draw[arr] (measure.south) -- ++(0,-3mm) -| (q.north);
\draw[arr] (q.west) -| node[above, pos=0.25] {yes} (cka.north);
\draw[arr] (q.east) -| node[above, pos=0.25] {no} (dh.north);
\draw[arr] (cka.south) -- (cka.south |- act.north);
\draw[arr] (dh.south) -- (dh.south |- act.north);
\end{tikzpicture}
\caption{Decision procedure for monitoring fine-tuning. The procedure follows from Sections~\ref{sec:why-sign} and~\ref{sec:why-early} and has been validated only on the configurations studied here.}
\label{fig:pipeline}
\end{figure}
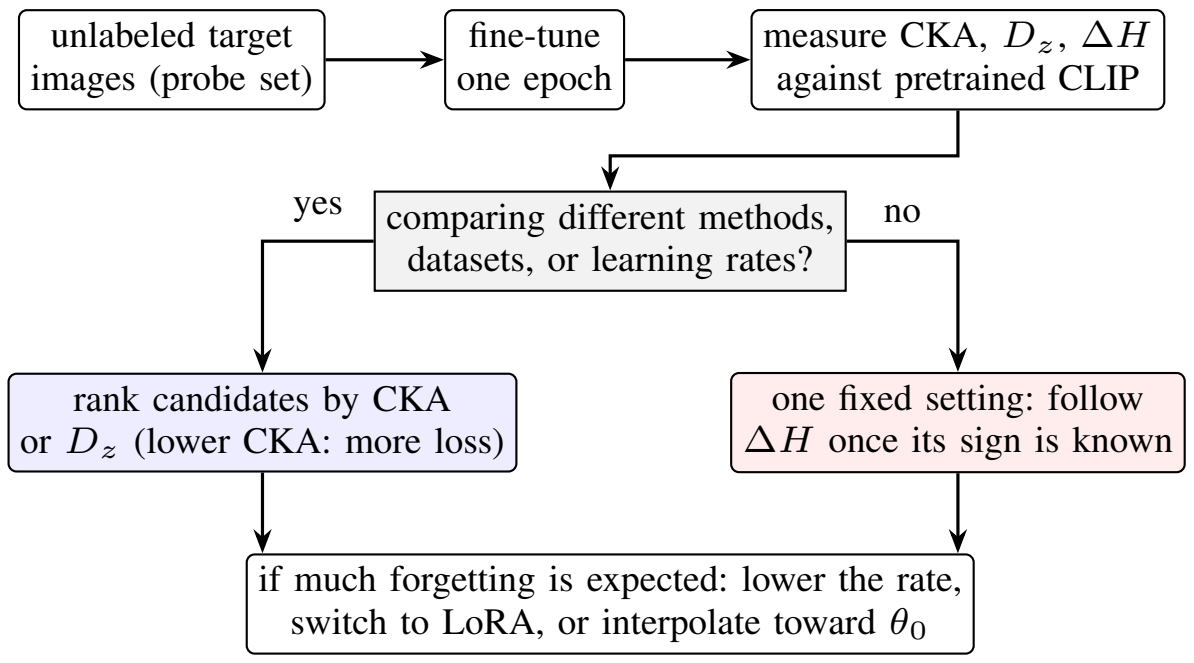

\subsection{The Accuracy and Retention Frontier}
No configuration dominates on both criteria. On EuroSAT, the linear probe preserves zero-shot accuracy but reaches only 85.1\% target accuracy, Full FT at $10^{-5}$ reaches 97.2\% but retains 41.2\%, and LoRA at $10^{-4}$ (95.2\%, 86.4\%) and tuning of the last block at $10^{-5}$ (88.6\%, 97.1\%) lie in between. The location of the update also matters: tuning only the last block at $10^{-4}$ changes the encoder weights more than Full FT at $3\cdot 10^{-5}$ yet retains 83.8\% rather than 22.3\%. Interpolating the encoder weights toward the pretrained model \cite{wortsman2022robust}, $\theta(\alpha)=(1-\alpha)\theta_0+\alpha\theta_1$, moves along this frontier: over three retrained seeds, $\alpha=0.5$ raises Full FT retention from 43.9\% to 90.8\% at a target accuracy of 94.5\%.

\section{Practical Guidelines}
Fig.~\ref{fig:pipeline} summarizes the procedure implied by these results. CKA is computed on a few hundred unlabeled target images after the first epoch, and candidates that differ in method, dataset, or learning rate are ranked by CKA or embedding drift. $\Delta H$ is appropriate only for monitoring a single configuration, once one or two runs have established which sign corresponds to stronger adaptation; $|\Delta H|$ should not be used to compare configurations. Weight change requires no images but can mislead: among DTD runs within two points of the best validation accuracy, it selects a run with 23.0\% retention when 86.2\% is attainable. When substantial forgetting is expected, the learning rate can be lowered, LoRA used instead of Full FT, or the weights interpolated toward the pretrained model.

\section{Limitations}
The study covers four target datasets (two held out), four learning rates, three seeds, six epochs, one model family (ViT-B at two patch sizes, with the held-out targets on ViT-B/32 only), and subsets of about 2{,}000 images for the secondary outcomes. Trends within groups are descriptive, and bootstrap intervals over 16 configurations are wide. All findings are correlational: interpolation shows that moving the weights toward the pretrained model restores retention, but not that attention or representation change causes forgetting. Attention entropy measures concentration rather than semantic content, and Fig.~\ref{fig:maps} shows single examples. Finally, CKA of the CLS token is closer to the zero-shot score than attention is; the results for early blocks and for the first epoch address this concern but do not eliminate it.

\section{Conclusion}
During CLIP fine-tuning, the change in attention entropy characterizes the magnitude of adaptation, and its sign is determined by the late blocks and by the task, so it tracks the learning rate within a configuration but cannot be compared across configurations. CKA with the pretrained model, computed on a few hundred unlabeled target images, predicts transfer degradation after the first epoch because the ordering of runs stabilizes early. Both findings hold on two held-out targets and two additional transfer benchmarks. Configurations should therefore be compared with CKA, and attention entropy reserved for monitoring a single configuration. Extending the evaluation to other model families and longer schedules is the natural next step.

\bibliographystyle{IEEEtran}
\bibliography{references}

\end{document}